\documentclass[conference]{IEEEtran}
\IEEEoverridecommandlockouts
\usepackage{cite}
\usepackage{amsmath,amssymb,amsfonts}
\usepackage{algorithmic}
\usepackage{graphicx}
\usepackage{textcomp}
\usepackage{xcolor}
\usepackage{booktabs}
\usepackage{enumitem}

\makeatletter
\newcommand{\linebreakand}{%
  \end{@IEEEauthorhalign}
  \hfill\mbox{}\par
  \mbox{}\hfill\begin{@IEEEauthorhalign}
}
\makeatother

\def\BibTeX{{\rm B\kern-.05em{\sc i\kern-.025em b}\kern-.08em
    T\kern-.1667em\lower.7ex\hbox{E}\kern-.125emX}}
    \usepackage{cite}
\usepackage[colorlinks=true, citecolor=blue, linkcolor=blue, urlcolor=blue]{hyperref}
\begin{document}

\title{State-Aware Virtual Patient Simulation for Training Psychiatrists in Challenging Clinical Scenarios\textsuperscript{*}
\thanks{\textsuperscript{*}This work is currently in progress, and the manuscript will be further refined and extended in subsequent versions.}
}

\author{
\IEEEauthorblockN{Harsh Mohta}
\IEEEauthorblockA{
\textit{Department of DSAI}\\
\textit{IIIT Bangalore}\\
Bengaluru, India\\
Harsh.Mohta@iiitb.ac.in
}
\and
\IEEEauthorblockN{C Chandrahas Reddy}
\IEEEauthorblockA{
\textit{Department of DSAI}\\
\textit{IIIT Bangalore}\\
Bengaluru, India\\
Chandrahas.CSVR@iiitb.ac.in
}
\and
\IEEEauthorblockN{Sunhith Reddy Bojja}
\IEEEauthorblockA{
\textit{Department of DSAI}\\
\textit{IIIT Bangalore}\\
Bengaluru, India\\
sunhith.reddy@iiitb.ac.in
}
\linebreakand
\IEEEauthorblockN{Manik Inder Singh Sethi}
\IEEEauthorblockA{
Asst. Professor, MD Psychiatry\\
\textit{National Institute of Mental Health}\\
\textit{and Neurosciences (NIMHANS)}\\
Bengaluru, India\\
manikindersingh@gmail.com
}
\and
\IEEEauthorblockN{Vinesh Gupta}
\IEEEauthorblockA{
\textit{Sanctuary Holistic Wellness}\\
\textit{Adj. Professor}\\
\textit{Royal Australian and NZ }\\
\textit{College of Psychiatrists}, Perth, Australia\\
Vingupt@gmail.com
}
\and
\IEEEauthorblockN{Raghuram Bharadwaj Diddigi}
\IEEEauthorblockA{
\textit{Department of DSAI}\\
\textit{IIIT Bangalore}\\
Bengaluru, India\\
raghuram.bharadwaj@iiitb.ac.in
}
}
\maketitle

\begin{abstract}
Psychiatrists must be prepared to manage patients exhibiting challenging psychological states, including suicidal ideation, hallucinations, and other high-risk behaviors. However, opportunities to safely practice such interactions are limited. While recent conversational agents can simulate patients, existing approaches largely rely on large language models (LLMs) to determine patient responses, making it difficult to consistently maintain psychological state, adapt to the clinician’s responses, and generate clinically plausible challenges that meaningfully test the clinician.
We present a state-aware virtual patient simulation pipeline for training psychiatrists in challenging clinical scenarios. The proposed system maintains an evolving patient state throughout the interaction and orchestrates patient behavior rather than leaving response generation entirely to an LLM. At each interaction, a learning-based behavioral strategy selector determines the type and level of challenge appropriate to the current patient and clinician context. The selected strategy, patient state, clinician response, interaction history, and relevant clinical knowledge are then provided as grounded context to an LLM to generate a semantically relevant and psychologically consistent patient response. Following each session, the system analyzes the interaction and provides feedback to the clinician on their communication and management of the virtual patient.
\end{abstract}


\section{Introduction}
The growing demand for mental health services worldwide has increased the need for well-trained psychiatric practitioners capable of handling patients with diverse and complex psychological conditions. Preparing clinicians to effectively manage such varied patient presentations requires not only a strong theoretical understanding of psychiatric conditions but also practical experience in handling real-world patient interactions. However, conventional training approaches, which rely heavily on theoretical instruction and supervised clinical exposure, can be time-consuming and may offer limited opportunities to repeatedly practice challenging clinical scenarios. The ability to respond appropriately to different patient behaviors, communicate effectively, and make sound clinical judgments is best developed through repeated practice in realistic settings. This motivates the development of automated training environments that simulate real-world patients, allowing psychiatrists to interact with virtual patients, practice their clinical and communication skills, learn from their responses, and progressively improve their preparedness for real-world clinical encounters.

Recent advances in Artificial Intelligence (AI), particularly Large Language Models (LLMs), present exciting opportunities for developing such automated training environments. Trained on vast and diverse corpora encompassing a wide range of human knowledge, including literature, philosophy, and medical sciences, LLMs possess considerable capabilities in understanding and generating human-like conversations. This makes them promising candidates for simulating realistic patient-clinician interactions. However, building a reliable virtual patient for psychiatric training is considerably more involved than simply prompting an LLM to act as a patient. Psychiatric interactions involve complex psychological states, nuanced behavioral responses, and evolving conversational dynamics that require consistency and contextual awareness throughout an interaction. A prompt-driven approach alone may not adequately capture these dependencies or ensure that the generated responses are clinically plausible and appropriately challenging. Ultimately, the effectiveness of an LLM-based system depends not merely on the language model itself, but also on the mechanisms that guide and constrain its behavior. Motivated by this, we propose an orchestrated simulation pipeline comprising dedicated modules for maintaining and updating patient psychological states, interpreting clinician responses, and learning appropriate patient behavioral strategies. These components work together to guide the LLM in generating contextually grounded patient responses, enabling a more realistic and adaptive training experience for psychiatric practitioners.

Towards this end, we develop a virtual patient simulation system that engages in sequential conversations with a clinician. In a real-world clinical interaction, a patient approaches the clinician with an underlying psychological state, which may evolve in response to the clinician's questions, interventions, and communication style. These evolving states influence how the patient perceives and responds to the clinician, including the nature, tone, and intensity of their responses. To emulate this process, our framework incorporates two key components: a state manager and a learning-based behavioral strategy selector. The state manager maintains and updates the patient's psychological state based on the ongoing interaction, while the learning algorithm identifies an appropriate behavioral strategy and its intensity, conditioned on the current patient state and conversational context. This behavioral guidance, together with the patient state, dialogue history, clinician's response, and relevant clinical information, is provided as structured context to the LLM, which generates the corresponding patient response. The generated response is subsequently evaluated to assess its behavioral consistency, alignment with the patient state, and appropriateness of the challenge posed to the clinician. The resulting reward is used to update the learning algorithm, enabling it to progressively improve its behavioral strategy selection. This interaction-evaluation-adaptation loop continues throughout the session, allowing the LLM to generate responses under explicit behavioral and contextual guidance rather than relying solely on unconstrained language generation. In essence, the learning algorithm determines how the patient should behave, while the LLM determines how that behavior should be expressed linguistically.

\section{Related Works}

\subsection{Background and Foundations}
Recent research highlights the importance of virtual patients in medical education to help clinicians train in adversarial situations. Hoo et al.~\cite{hoo2026use} review the use of different types of virtual patients in psychiatry training and discuss how they are being used in medical education. Zeng et al.~\cite{zeng2025embracing} review recent trends and challenges of LLM-based virtual patients. Their review highlights the need for more standardized evaluation methods and assessment tools to ensure the quality of simulated patients.

Zhang et al.~\cite{zhang2026human} examine the differences between LLM-generated standardized patients and human standardized patients using multiple evaluation criteria. Hoang et al.~\cite{hoang2026psi} evaluate depression patient simulators at the turn, dialogue, and population levels to assess their clinical realism and consistency. Similarly, Emerson et al.~\cite{emerson2025automated} propose evaluation metrics for human and simulated standardized patients using two LLM as a judge approaches that align with human evaluators. Together, these studies provide different ways to evaluate how realistic and consistent simulated patients are.
 
\subsection{Steering simulated patients and personas}

The most common problems faced when simulating patients with LLMs are oversharing and insufficient diversity in patient behavior.

Wu et al.~\cite{wu2026emopatient} address the problem of patient's emotions being static in patient simulators. They proposes an Emotion Director that calculates the patient's emotional state using the latest doctor-patient exchange and generates control signals at every turn to guide the next response. However, it does not explicitly consider how a patient's willingness to disclose information changes in the conversation.

Schlager et al.~\cite{schlager2026patientspersonalityrealisticpatient} represent patient personality using the six HEXACO dimensions and employ a latent patient state to control conversational style, cooperativeness, and selective disclosure. They use repeated and targeted questioning as an indication for progressively revealing medical facts and assign different behavioral roles to patients based on their personalities. However, this approach does not explicitly consider how a patient's willingness to disclose information changes in response to the doctor's behavior and the trust developed during the conversation.

\subsection{Mental-health-specific patient simulation}

Xu et al.~\cite{xu-etal-2026-puppet} propose a neural-symbolic VSP for mental health that uses expert-defined rules to connect clinician interventions with changes in patient mental states. It uses an Observe–Think–Behave architecture to make simulated patients respond to clinical interventions in a controlled manner. However, covering diverse patient profiles and conversational scenarios might require additional rule design, as it relies on expert-defined state-transition rules.

Sabour et al.~\cite{sabour-etal-2026-patientact} present a theory-based approach to simulating mental health clients for therapist training. They model patient emotions and behaviors and use trust thresholds to control when sensitive information can be revealed. However, our work focuses on psychiatric consultations using patient profiles based on interview transcripts. We combine trust thresholds with conditions related to the clinician's behavior and the current situation to control information disclosure. We also use a learning algorithm to learn which patient behaviors are suitable in different consultation contexts and evaluate response quality, information leaks, and behavioral diversity across full consultations.

\subsection{Persona-based and State-aware Patient Simulation}

Kyung et al.~\cite{NEURIPS2025_24945e3b} propose a patient simulator that generates diverse patient personas for doctor-patient interactions. It uses four dimensions (personality, language proficiency, medical-history recall, and cognitive confusion) to simulate different patient characteristics. However, it focuses on creating patients with different characteristics rather than learning from feedback to select patient responses as the conversation progresses.

\subsection{LLM steering, contextual bandits, and human feedback}

Deng et al.~\cite{deng2023prompting} propose Proactive Chain-of-Thought, a prompting method that prompts an LLM to select an appropriate strategy before generating a response. However, the authors found that the results are affected by the choice of prompts and are weak compared to fine-tuned methods. They also noted that simple prompting may not be enough for decision-making tasks in real-world applications.

Deng et al.~\cite{ICLR2024_29e8437d} propose Plug-and-Play Policy Planner (PPDPP), which uses a trainable policy planner to choose dialogue strategies for an LLM-based dialogue agent. The planner is trained using supervised learning on human-annotated data as well as reinforcement learning using feedback from an AI evaluator.
%
However, the paper limits its experiments to eight turns per conversation, whereas psychiatric consultations may involve longer interactions in which patient behavior and trust change over time.

Previous work has explored different ways to control simulated patient emotions, personalities, behavior, and information disclosure. However, selecting appropriate patient behaviors throughout longer psychiatric consultations remains an important challenge. Our work addresses this by combining patient state tracking with a learning-based approach that selects patient behavior according to the current patient state and conversation context. We also compare different behavior selection methods across extended consultations.

To address these challenges, we implemented the following:

\begin{itemize}
    \item \textbf{Adaptive patient behavior selection.}
    We use a contextual bandit approach that selects the patient's behavior based on updated patient state and conversation context.

    \item \textbf{Extended consultation evaluation.}
    We evaluate the simulator across longer consultations with different clinician behaviors and their interaction styles.

    \item \textbf{Comparative evaluation of behavior selection.}
    We compare LLM-based and learning-based behavior selection in terms of response quality, hidden-information leaks, and diversity of patient behavior.
    
\end{itemize}

\section{Methodology}
\subsection{System Overview}
\begin{figure*}[h!]
    \centering
    \includegraphics[width=0.9\textwidth]{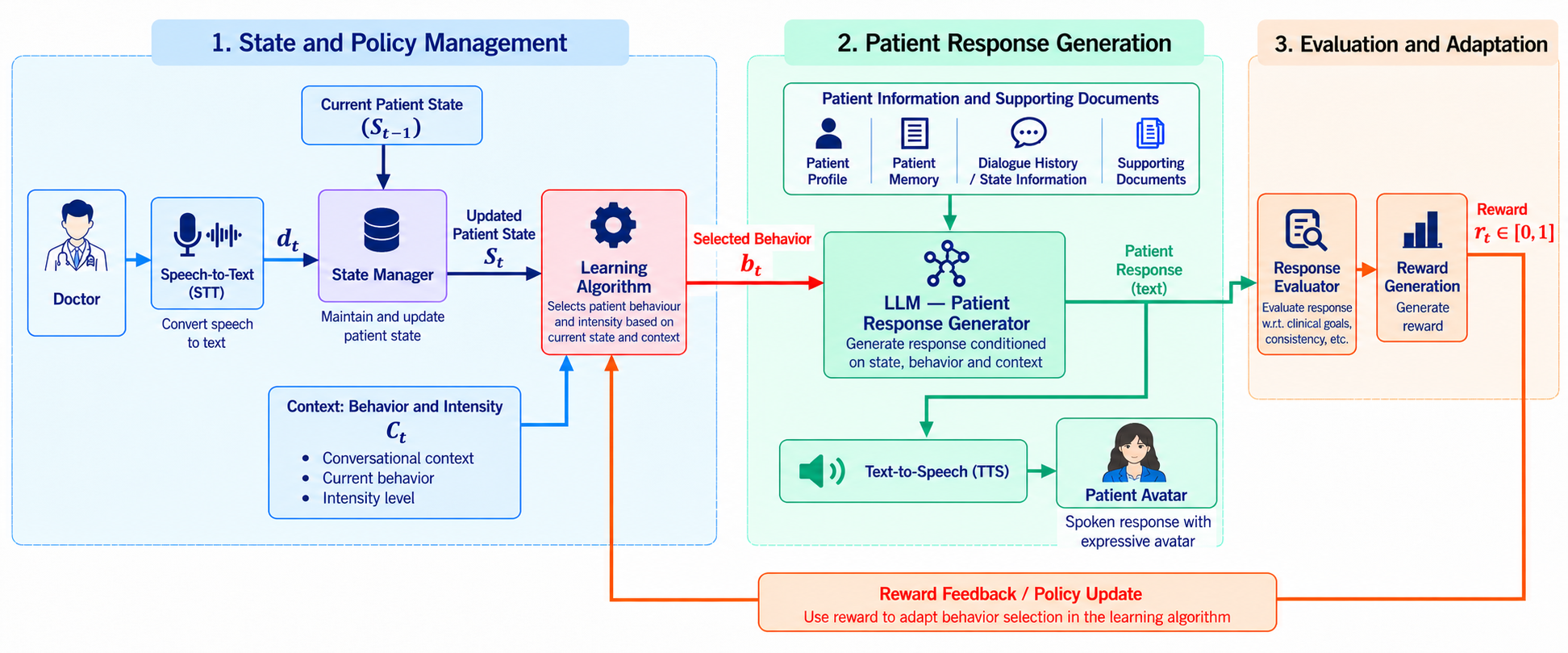}
    \caption{System architecture of the proposed approach.}
    \label{fig:architecture}
\end{figure*}

We now propose an LLM-based end-to-end pipeline for simulating patients in psychiatric consultations. As shown in Fig.~\ref{fig:architecture}, our pipeline consists of three main stages: \begin{enumerate}[label=(\roman*)]
  \item State and Policy Management
  \item Patient Response Generation
  \item Response Evaluation and Adaptation
\end{enumerate}

During each conversation turn, the clinician's utterance is converted to text and given to the state manager, which maintains the evolving state of the patient. Based on the updated state and context (as shown in Fig.~\ref{fig:architecture}), the learning algorithm then decides a behavioral strategy. The selected strategy guides the response LLM, which uses the patient's profile, conversation history, and other relevant information to produce a contextually appropriate response.

The generated response is then evaluated using predefined criteria, and the resulting reward is used to train the learning algorithm. The generated response is also converted into speech using text-to-speech (TTS) and presented through a patient avatar, providing an interactive interface for psychiatric consultations.

\subsection{Patient Representation and State Management}

\subsubsection{Patient Profile}
For each simulated patient, we curate a patient profile, which is a structured template that contains information about the patient's identity, clinical case, psychological model, speech profile, hidden information, and goals. The psychological model describes the patient's beliefs, fears, and their interpretation biases. The speech profile describes how the patient speaks, mentioning their vocabulary, language, and characteristic expressions. We include hidden information in the patient profile, which allows the system to control when specific information becomes available during the conversation.

\subsubsection{Patient State and State Manager}
The patient state consists of different aspects of the patient's current condition, such as their emotional state and stance toward the clinician, which allows the system to track changes in the patient's behavior as the conversation progresses.

The state manager takes the latest response of the clinician, relevant information from the patient profile, current patient state, and conversation history as inputs to update the patient's state. We use predefined categories to represent the different aspects of patient state. For example, the patient's feelings are grouped into seven categories, while their stance toward the clinician is grouped into four categories. This predefined representation helps to maintain consistency in tracking the state and behavior in conversations.

\subsubsection{Appointment Memory}

Psychiatric consultations can involve long conversations, so passing the entire conversation history to the response LLM at every turn would be impractical. Instead, the response LLM receives the last 12 exchanges and a short memory of what the patient has already told the clinician. The state manager also receives this memory. This helps the patient stay consistent and avoid repeating information as if it were being shared for the first time.

The memory is organized into broad topics, such as family and daily life, with a short summary for each topic. It records information the patient has shared during the consultation, including details not included in the original patient profile, as well as questions the patient refused to answer or avoided. A separate LLM call updates the memory using the patient profile, the existing memory, and the latest exchanges. Any hidden information that has not yet been revealed is removed from the profile before this update, preventing it from being added to the memory. The LLM returns only new or changed information, which is then added to the existing memory. All other entries remain unchanged.

The memory is updated in the background, so the patient's response does not have to wait for the update to finish. This means the memory may be slightly behind the conversation, but the most recent exchanges are always provided directly to the response LLM.

\subsection{Behavior Representation and Selection}

Initially, our state manager uses an LLM to select the patient's behavioral response based on the patient state and the context. However, using an LLM to select specific behavior adds computational cost and can increase response latency. To overcome this challenge, we introduce a learning-based behavior-selection algorithm that is responsible for selecting the patient's behavior. This method reduces the decision-making task burden on the state manager and allows it to focus on tracking changes in the patient's state.

\subsubsection{Behavior Representation}
We represent patient behavior using four dimensions: feelings, stance toward the clinician, kind of reply, and behavior strength. Each dimension captures a different aspect of the intended patient response. A behavior representation at turn $t$ is represented as

\begin{equation}
    a_t = (f_t, s_t, k_t, g_t), 
\end{equation}

where $f_t$, $s_t$, $k_t$, and $g_t$ represent the patient's feeling, stance toward the clinician, kind of reply, and behavior strength, respectively.

Each dimension consists of a predefined set of categories: seven possible feelings, four stances, eight kinds of replies, and three strength levels. Together, they form the set of behavior configurations
\begin{equation}
    \mathcal{A} = \mathcal{F} \times \mathcal{S}
    \times \mathcal{K} \times \mathcal{G}.
\end{equation}

The total number of possible behavior configurations is therefore

\begin{equation}
    |\mathcal{A}| = 7 \times 4 \times 8 \times 3 = 672.
\end{equation}

\subsubsection{Contextual Reward Estimation}

We consider the behavior selection problem as a contextual multi-armed bandit task and use a LinUCB-based algorithm~\cite{li2010contextual}, where each behavioral configuration is an arm. At each conversation turn, the algorithm receives a context vector $c_t$ describing the current conversation context and patient state.

We represent the relationship between the context and a candidate behavior configuration using a joint feature vector $\phi(c_t, a)$. Following the linear reward model of LinUCB~\cite{li2010contextual}, we estimate its expected reward as

\begin{equation}
    \hat{r}(c_t,a) = \theta^\top \phi(c_t,a),
\end{equation}

where $\theta$ represents the learned model parameters. Inspired by the hybrid model of LinUCB~\cite{li2010contextual}, which shares parameters across arms, we expand this estimate as

\begin{equation}
    \begin{aligned}
        \hat{r}(c_t,a) ={}&
        \beta^\top c_t
        + w_f^\top c_t
        + w_s^\top c_t
        + w_k^\top c_t\\
        &+ v_g^\top c_{S,t}
        + p_{fs}+p_{fk}+p_{sk}+p_{kg}\\
        &+\sum_{m\in\{f,s,k,g\}}
        \eta_m\mathbb{I}[a_m=a_{t-1,m}].
    \end{aligned}
\end{equation}

Here, $\beta$ represents the context-dependent baseline reward, while $w_f$, $w_s$, and $w_k$ capture the contributions of feelings, stance, and kind of reply, respectively. 
The term $v_g$ captures the contribution of behavior strength using a subset of context features $c_{S,t}$. The terms $p_{fs}$, $p_{fk}$, $p_{sk}$, and $p_{kg}$ capture the interactions between different behavioral dimensions. 
The final term captures the contribution of maintaining the same behavior dimension as the previous turn, which encourages consistency across consecutive responses.

\subsubsection{Behavior Selection and Policy Learning}

The algorithm uses the rewards obtained from the response evaluation process to learn the model parameters. We use ridge regression~\cite{vanwieringen2026lecturenotesridgeregression} to estimate these parameters based on the collected training samples.
The parameter estimates are given by~\cite{vanwieringen2026lecturenotesridgeregression, li2010contextual}

\begin{equation}
    A = \lambda I + \sum_{i=1}^{n}\phi_i\phi_i^\top,
\end{equation}

\begin{equation}
    b = \sum_{i=1}^{n}R_i\phi_i,
    \qquad
    \hat{\theta}=A^{-1}b,
\end{equation}

where $\phi_i$ represents the joint feature vector for the $i$-th training sample, $R_i$ is the corresponding reward obtained from the evaluator, $\lambda$ is the regularization parameter, and $I$ is the identity matrix.

Additionally, we compute the uncertainty associated with each candidate behavior configuration as~\cite{li2010contextual}

\begin{equation}
    \omega(c_t,a) =
    \sqrt{\phi(c_t,a)^\top A^{-1}\phi(c_t,a)}.
\end{equation}

During the learning phase, we use uncertainty as an exploration bonus to encourage the algorithm to try behavior configurations for which it has less information. The exploration score is calculated as~\cite{li2010contextual}

\begin{equation}
    S_{\text{explore}}(c_t,a)
    =
    \hat{r}(c_t,a)+\alpha\omega(c_t,a).
\end{equation}

The behavior configuration is then sampled using a softmax policy~\cite{sutton2018reinforcement}

\begin{equation}
    \pi(a\mid c_t)=
    \frac{\exp(S_{\text{explore}}(c_t,a)/T)}
    {\sum_{a'\in\mathcal{A}}
    \exp(S_{\text{explore}}(c_t,a')/T)}.
\end{equation}

The selected behavior is then passed to the patient response LLM, which generates a reply while remaining consistent with the selected behavior and patient profile. 

The model can be trained using the previously collected consultation data. It can further learn online during the simulated consultations by updating parameters based on human feedback from clinicians on the generated patient responses. This allows the algorithm to adapt its behavior-selection strategy as more feedback becomes available.

\subsection{Information Disclosure Control}
As discussed in previous sections, the patient profile contains hidden information that is associated with specific opening conditions and trust thresholds. Only when the specific opening condition is satisfied, and the patient's trust reaches the required threshold, is the hidden fact available for disclosure.

Based on the clinician's behavior, trust is updated. It increases when the clinician is considerate, remains the same when the clinician is neutral, and decreases when the clinician is dismissive. This mechanism helps prevent the patient from disclosing information easily and encourages the clinician to gain the patient's trust and ask the right questions.

\subsection{LLM-Based Patient Response Generation}

The response LLM generates the patient's response using the selected behavior, patient profile, appointment memory, recent conversation, and supporting documents. The selected behavior guides the response LLM to generate how the patient expresses themselves according to the situation while being consistent with the patient's profile.

The response LLM receives only the hidden information made available by the disclosure-control mechanism, which ensures the patient LLM cannot directly access the hidden information. This allows information to be revealed progressively as the conversation develops. The generated response is returned as plain text and can be converted into speech using TTS.

\subsection{Response Evaluation and Reward Computation}
The generated response is evaluated using a set of predefined checks and an LLM-based evaluator. The evaluation measures how well the response fits the patient, the current interaction, natural speech, and consistency with the conversation.

The scores for each criterion are then combined to produce a reward that provides feedback for training the learning algorithm. We also designed a few other checks to identify responses that reveal hidden information when not necessary or fail to follow the selected behavior.

\subsection{Trainee Assessment and Feedback}
After each consultation, the trainee needs to write a short case summary, list up to 3 possible diagnoses which they found, and provide a statement about the risk. An LLM then reviews the consultation transcript and the trainee's formulation using a rubric covering seven areas (communication and rapport, interview structure, history-taking, mental state examination, risk assessment, diagnostic reasoning, and professionalism). The rubric is based on the Master Interview Rating Scale~\cite{mirs2009}, the RCPsych CASC syllabus~\cite{rcpsych2023casc}, and the ABPN Clinical Skills Evaluation~\cite{abpn_cse}. For each rubric item, the LLM gives a verdict and we make the LLM to cite evidence from the transcript. It also assigns each area a score from 1 to 5. Any case information that remained hidden at the end of the consultation is removed before the LLM receives the case.

We use a separate LLM call to give the trainee feedback on their performance. It highlights what they did well, identifies the two or three most important things they could improve, and suggests how to work on them. If any area needs improvement, it explains why it matters in clinical practice and gives an example of how the trainee could phrase something differently.

\bibliographystyle{IEEEtran}
\bibliography{references}
\end{document}